%% file: paper_icml_rev2_arxiv.tex
\documentclass{article}

\usepackage{microtype}
\usepackage{graphicx}
\usepackage{booktabs}
\usepackage{enumitem}
\usepackage{amsmath}
\usepackage{amssymb}
\usepackage{mathtools}
\usepackage{tikz}
\usetikzlibrary{arrows.meta,positioning}
\usepackage{hyperref}
\usepackage{xcolor}

\usepackage[preprint]{icml2026}

\makeatletter
\def\toptitlebar{\vskip 0.18in}
\def\bottomtitlebar{\vskip 0.18in}
\makeatother


\newcommand{\W}{\mathcal{W}}
\newcommand{\Q}{\mathcal{Q}}
\newcommand{\R}{\mathcal{R}}
\newcommand{\B}{\mathcal{B}}
\newcommand{\D}{\Delta}
\newcommand{\E}{\mathbb{E}}
\newcommand{\Min}{M_{\mathrm{in}}}
\newcommand{\Mout}{M_{\mathrm{out}}}
\newcommand{\Bm}{B_{\mathrm{micro}}}
\newcommand{\Mref}{M^{\mathrm{ref}}}
\newcommand{\MT}{M^{\mathrm{T}}}
\newcommand{\MF}{M^{\mathrm{F}}}

\icmltitlerunning{Macro Control and Capability Evaluation of Black-Box World Models}
\hypersetup{pdftitle={Macro Control and Capability Evaluation of Black-Box World Models: A Two-Regime Framework for LLM-Agent Markets},
  pdfauthor={Zhiyang Niu, Chuanjie Li, Huayuan Shen, Mengmeng Zhao, Jiajia Zhang}}

\begin{document}

\twocolumn[
  \icmltitle{Macro Control and Capability Evaluation of Black-Box World Models:\\
  A Two-Regime Framework for LLM-Agent Markets}

  \begin{icmlauthorlist}
    \icmlauthor{Zhiyang Niu}{hit}
    \icmlauthor{Chuanjie Li}{ntu}
    \icmlauthor{Huayuan Shen}{hit}
    \icmlauthor{Mengmeng Zhao}{pingan}
    \icmlauthor{Jiajia Zhang}{hit}
  \end{icmlauthorlist}

  \icmlaffiliation{hit}{Harbin Institute of Technology, Shenzhen, China}
  \icmlaffiliation{ntu}{Nanyang Technological University, Singapore}
  \icmlaffiliation{pingan}{Pingan Group, China}

  \icmlcorrespondingauthor{Jiajia Zhang}{zhangjiajia@hit.edu.cn}

  \icmlkeywords{world models, LLM agents, financial markets, macro control, black-box evaluation, counterfactual evaluation}

  \vskip 0.3in
]

\printAffiliationsAndNotice{}

\begin{abstract}
Large language model (LLM) agents can be organized into multi-agent markets that act as world models, mapping macro information to emergent macro dynamics while the evaluator observes only macro inputs and outputs. This paper separates a model's capability into two stages that a single aggregate score cannot
separate: \emph{information simulation}, from macro information to micro behavior,
and \emph{macro emergence}, from micro behavior to the macro outcome. We introduce a soft macro-control method that announces a public reference path day by day without touching individual decisions, orders, or matching, and a two-regime evaluation framework that measures both stages from macro observations
alone. In the reference-only regime the symmetric true and flipped references share a common level $R$, which measures realization under the announced reference, while their paired gap $B$ measures structural bias; in the information-injection regime external information dilutes the reference, so only the true--flipped gap $\Delta$ remains interpretable. On five single-index LLM-agent markets, announcing a target path shifts the realized path toward the announced target on both
announced references under both information conditions. Relative to the no-reference path, the tracking score is higher by $0.306$ and $0.415$
without news and by $0.183$ and $0.329$ with news; within the stock/cash scan structural bias is lowest at 70/30 ($B=0.042$); realization peaks at a 3:7 technical share ($R=0.894$ versus $0.358$ for technical-only); and $\Delta$ is $0.0341$ with fundamental data and $0.0487$ with news. Both stages use the same two observable scores, each read in the regime where it is interpretable.
\end{abstract}

\section{Introduction}
\label{sec:intro}

Large language model (LLM) agents can be organized into a multi-agent market and used as a world model: the model is given macro-level conditions, agents trade and communicate, and macro-level prices emerge from their interaction \citep{park2023generative,yang2025twinmarket}. The object of interest is not a single fitted price path but the transformation chain
\begin{equation}
\Min \;\longrightarrow\; \Bm \;\longrightarrow\; \Mout,
\label{eq:pipeline}
\end{equation}
where $\Min$ is the macro information available to the agents (prices, news,
fundamentals), $\Bm$ is micro-level behavior, and $\Mout$ is the resulting macro
phenomenon. Macro control in this paper is the addition of \emph{one further
information item}: a public reference level, presented as an external reference datum
rather than as a trading instruction. The evaluator observes the two ends of this chain; the middle is the model's internal implementation.

In a real system micro behavior is often observable, but it belongs to the model, not
to the evaluator: throughout, \emph{black box} refers to the evaluation interface,
where the evaluator may query only $(\Min,\Mout)$ pairs. The restriction makes
different models comparable without access to their internals.

Capability is not one quantity: the chain of Eq.~\eqref{eq:pipeline} has two stages, and each
corresponds to a distinct ability, \emph{information simulation} from macro information to
micro behavior and \emph{macro emergence} from that behavior to the macro outcome. A model can fail at either stage, and the failures need different measurements: agents may use
the information correctly while the market structure cannot support the resulting macro
emergence (realization); the simulation may be hard-coded in that structure, so that a different
announced path cannot be realized (structural bias); or the injected information may leave the
output unchanged (information effectiveness). One aggregate score over a realized path cannot
tell these cases apart. The framework below separates them by manipulating what information the
model receives, and measures all three from macro observations alone.

A common evaluation practice is to compare generated macro paths with historical paths and ask whether stylized statistics are reproduced \citep{lebaron2006agent,hommes2006heterogeneous,yang2025twinmarket,piao2026marketsim}. Prior work on LLM-agent markets has largely evaluated agents as traders, reporting returns, risk, or decision quality \citep{yu2023finmem,zhang2024finagent,li2025investorbench}. Both describe a simulation, but neither separates the three properties above: a similarity to
history uses one realized path, so it cannot show how the model treats a different announced
path or whether injected information changes its behavior, and trading metrics say nothing
about the macro map.

A second difficulty is that any absolute path score is comparable only
within a fixed information condition, because injecting external information
changes the task itself---the announced reference must compete with news or
fundamentals, so it can be diluted and the score can fall even when the model's
behavior is unchanged. The framework therefore interprets absolute scores only in
a reference-only regime and switches to the true--flipped difference $\Delta$ once
information is injected.

Turning news on, for example, lowers absolute TSR/FSR from $0.9180/0.8839$ to
$0.8307/0.7820$ yet raises $\Delta$ from $0.0341$ to $0.0487$, so a naive reading
of the absolute scores would reach the opposite conclusion.

\paragraph{Contributions.}
\begin{enumerate}[leftmargin=*]
  \item \textbf{A soft macro-control method.} We expose a public macro reference path to a black-box world model while leaving individual decisions, orders, and matching untouched; control effectiveness is the increase in the true-target tracking score relative to a
  no-reference arm, and it measures the influence of the announcement rather than a
  capability of the model.
  \item \textbf{A two-regime black-box evaluation framework.} We separate a
  reference-only regime, in which absolute scores are interpretable, from an
  information-injection regime, in which only the true--flipped difference is interpretable.
  \item \textbf{A capability decomposition.} Realization capability $R$ and
  structural bias $B$ are measured in the reference-only regime, and information
  effectiveness $\Delta$ in the injection regime, reusing the same two scores.
  \item \textbf{A calibrated empirical study.} On five single-index LLM-agent
  markets we study control effectiveness, structural constraints, agent
  composition, fundamental information and news, identify a low-bias operating
  point and a non-monotonic share--realization relation, and find positive
  information effects for fundamental data and news; the scores are calibrated
  against null baselines and alternative definitions.
\end{enumerate}

\section{Related Work}
\label{sec:related}

\emph{Generative social simulation.} Generative agents with memory, reflection and
planning sustain coherent behavior over long horizons \citep{park2023generative},
and related prototypes populate social-computing systems \citep{park2022social};
demographically conditioned models reproduce survey response distributions
\citep{argyle2023out} and act as implicit models of economic agents
\citep{horton2023silicon}; EconAgent simulates macroeconomic activity
\citep{li2024econagent}; AgentSociety scales LLM societies to large populations
\citep{piao2025agentsociety}; and generative agent-based modeling couples such
agents with mechanistic simulation \citep{ghaffarzadegan2024gabm}. 

\emph{LLM agents in financial markets.} LLM trading agents add layered memory
\citep{yu2023finmem}, an idea already present in earlier character-based trading
agents \citep{li2023tradinggpt}, multimodal tool use \citep{zhang2024finagent},
and multi-agent debate \citep{yu2024fincon}, and are compared on trading
performance or benchmark scores \citep{li2025investorbench}; language models are
also used as financial information processors that score news and predict
subsequent returns \citep{lopezlira2026chatgpt}. Multi-agent market simulators
instead model the market itself: StockAgent embeds LLM agents in a trading
environment \citep{zhang2024stockagent}, TwinMarket couples beliefs, a social
network and an order book \citep{yang2025twinmarket}, and MarketSim places 15k
generative participants in a continuous double auction \citep{piao2026marketsim}.
These works evaluate a simulator by how closely its statistics match history, not
by how it responds to controlled changes in the public information.

\emph{World models, evaluation and control.} Learned world models support planning
and control \citep{ha2018world,schrittwieser2020muzero,hafner2025dreamer}, and
interpretability probes recover internal representations of the environment
\citep{li2023othello,gurnee2024language}; agent benchmarks measure task success in
controlled environments \citep{liu2024agentbench,kapoor2024aiagents}, evaluation
methodology contributes behavioral tests, multi-scenario benchmarks and audits
\citep{ribeiro2020checklist,liang2023helm,reuel2024betterbench}, and validation
practice separates calibration from validation for simulators
\citep{windrum2007empirical}; counterfactual semantics supplies the language for
alternative inputs \citep{pearl2009causality}, while activation steering
intervenes inside the model \citep{turner2023steering}. Agent-based computational
finance studies how heterogeneous micro rules generate macro regularities
\citep{lebaron2006agent,hommes2006heterogeneous,tesfatsion2006ace}; switching
rules produce complex price dynamics \citep{brock1998heterogeneous}, interacting
agents reproduce scaling and criticality \citep{lux1999scaling}, and the tradition
argues for interacting-agent models
\citep{lebaron2008macro,farmer2009economy,axtell2025agent}; policy learning has
also been studied inside simulated economies \citep{zheng2022aieconomist}. This paper instead treats the market as a black-box world model, controls it only
through public macro information, and separates the information regimes.

\section{Problem Formulation and Framework Overview}
\label{sec:problem}

\subsection{Black-Box World Models}
\label{sec:problem:model}

Let $\W$ be a market world model. The macro input is public information, which we represent as a path $M=(M_1,\ldots,M_T)$ over $T$ trading days, and the corresponding macro output is a simulated path $P=(P_1,\ldots,P_T)$. At the evaluation interface the model is the map
\begin{equation}
P = \W(M).
\label{eq:map}
\end{equation}
Because the two stages of Eq.~\eqref{eq:pipeline} can fail independently, evaluation must
separate realization, structural bias and information effectiveness rather than summarize all
three with one tracking score.

\subsection{Capability Decomposition}
\label{sec:problem:decomp}

The framework distinguishes three properties of a black-box world model.

\emph{Realization capability.} Can the model realize a given macro path at all, including paths that did not occur
historically? Both stages must work: the agents must act on the information they receive, and
the micro structure must aggregate those actions into that macro path. A model that ignores its
input, or whose structure is too narrow to support the path, fails, even if its output happens to
resemble one particular history.

\emph{Structural bias.} Does the model treat two counterfactual announcements of the same form differently when no effective external information is available? A systematic difference indicates that the market structure, trading constraints, or path construction introduces a directional preference.

\emph{Information effectiveness.} When external information is injected, does the model use it to distinguish a true announcement from a mirrored one? This is the only property that can be attributed to an information channel, and it must be measured differently from the two properties above.
Realization capability and structural bias concern macro emergence, and information effectiveness
concerns information simulation.

\subsection{Two Regimes}
\label{sec:problem:regimes}

A central design choice of this paper is that the interpretation of a path score depends on what information the model receives.

In the \emph{reference-only regime}, no external news or fundamental data is injected. The model receives a true reference path and a flipped reference path in separate arms. Because the two references are symmetric, the true-signal realization score $\mathrm{TSR}$ and the flipped-signal realization score $\mathrm{FSR}$ should be approximately equal: their common level measures realization capability, and their paired gap measures structural bias.

In the \emph{information-injection regime}, external information is added. The information can change micro behavior and can also dilute or override the injected macro reference. Absolute TSR and FSR then no longer admit a comparable realization interpretation; the only interpretable quantity is their difference $\D=\mathrm{TSR}-\mathrm{FSR}$.

The two regimes differ in the surrounding information environment---whether
news and fundamental data are present---not in how the reference is delivered.

\begin{figure}[t]
  \centering
  \input{figs/fig1_framework_tikz}
  \caption{Overview of the framework. Top: the evaluator sees only the black-box map $\W$ from the public macro information to a macro output path; micro behavior is internal and unobserved. The evaluator announces the public macro reference level at the start of each trading day. (a) In the reference-only regime the model is run on a true reference and on its flipped mirror, so the common level $R$ measures realization capability and the root-mean-square paired gap $B$ measures structural bias. (b) In the information-injection regime external information is added; absolute scores are no longer comparable, and only $\D=\mathrm{TSR}-\mathrm{FSR}$ is interpreted.}
  \label{fig:framework}
\end{figure}
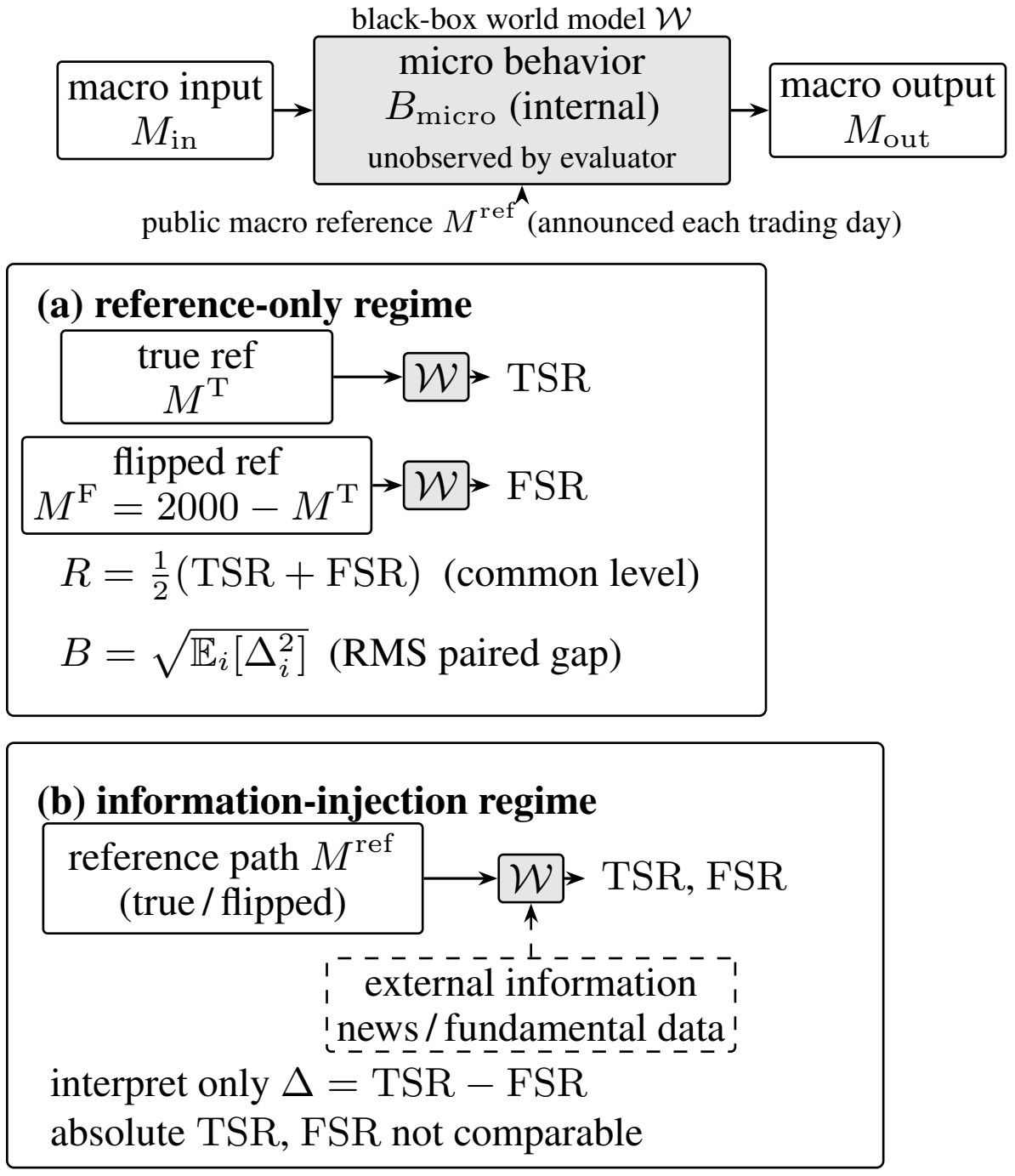

The separation is part of the framework, not a post-hoc analysis choice: applying a reference-only metric to an information-injection experiment, or the reverse, produces misleading conclusions. Figure~\ref{fig:framework} summarizes the framework; Appendix~\ref{app:notation} collects the notation used in the remainder of the paper.

\section{Macro Control via a Public Reference}
\label{sec:control}

\subsection{Why Control Must Be Soft}
\label{sec:control:soft}

The macro output of a world model is generated by micro behavior. Directly forcing the output is not meaningful: writing the desired price into the system is an assumption, not a capability. Intervening in individual decisions, orders, or matching is also undesirable,
because it destroys the self-organization that makes the model a world model. The control problem is therefore to steer the macro output using only public information, without changing the model's internal implementation.

\subsection{Public Reference Protocol}
\label{sec:control:protocol}

At the beginning of each trading day $t$ the evaluator announces the public macro reference level $\Mref_t$ for that day. Over the horizon, these daily announcements define a public macro reference path
\begin{equation}
\Mref=\{\Mref_1,\ldots,\Mref_T\}.
\label{eq:ref}
\end{equation}
In this implementation, the level announced for day $t$ is the normalized target level for that day, which is derived from that day's close (or from its flipped counterpart). The announcement therefore specifies the outcome to be realized, not a forecast. The
reference is simply the target path chosen by the evaluator: this implementation uses the realized path, and its mirrored counterpart for
the flipped arm, but any target path can be announced. Choosing the realized path is a
design decision, not a claim that the announcement improves fidelity to the data. Only the current day's value is revealed; the agents do not observe the rest of the path. Every agent receives the same daily signal, and whether and how the signal is used is decided inside the model.
Agents are never told to track the reference: nothing distinguishes it from the
other information they receive except its content, and whether they use it is what
the experiment measures. The method has three properties: it is macro-level only (the information content is a path, not an order or an agent-specific instruction), it involves no micro intervention (strategies, beliefs, order sizes, and matching rules are untouched), and it is public (all agents receive the same signal). It is therefore a macro-level \emph{soft} control: the evaluator controls what information is supplied, not the internal mechanism.

The reference and no-reference arms use identical prompts except for the reference
block, so the announcement is the only difference.

\subsection{Control Effectiveness}
\label{sec:control:gain}

For a target path $M^\star$, control effectiveness is the increase in the tracking score when the reference is present:
\begin{equation}
\begin{split}
\mathrm{Gain}(M^\star)
&=
\Q\bigl(\W_{\mathrm{ref}}(M^\star),M^\star\bigr)\\
&\quad-
\Q\bigl(\W_{\mathrm{noref}},M^\star\bigr),
\end{split}
\label{eq:gain}
\end{equation}
where $\W_{\mathrm{ref}}$ is the model with the public reference and $\W_{\mathrm{noref}}$ is the same model without it. A positive gain means that the public reference moves the macro output toward the
announced target. Gain is a tracking quantity, not a capability estimate: with no
announcement the same market sits at the zero-information level on the true target
(Section~\ref{sec:results:validation}), so the gain is attributable to the announcement.

Control effectiveness and information use are different questions. A model can obtain a
positive gain by echoing the announced path without using genuine information; conversely,
it can use information well and still track a given path poorly, if its structure cannot
realize that path. The gain alone therefore cannot certify information use, which is why the
control experiment is complemented by the framework of Section~\ref{sec:evaluation}.

We evaluate the gain on the true and flipped references under two information conditions---news off and news on---which separates the experiment into four groups. Every arm-level comparison has a positive gain (Section~\ref{sec:results:control}).

\section{Evaluation Framework}
\label{sec:evaluation}

\subsection{Path Realization Score}
\label{sec:evaluation:q}

For a realized path $P$ and a target path $M$, define the daily relative error in percent
\begin{equation}
e_t=\left|\frac{P_t}{M_t}-1\right|\times 100 .
\label{eq:err}
\end{equation}

\paragraph{Level.}
Path errors accumulate over the evaluation horizon, so we normalize the mean absolute error by a path-level volatility scale,
\begin{align}
r_t &= 100\bigl(\log M_t-\log M_{t-1}\bigr),\\
\sigma_M &= \operatorname{std}(r_t),\\
\mathrm{NMAE} &= \frac{\operatorname{mean}(e_t)}{\sigma_M\sqrt{T}},\\
\mathrm{Level} &= \frac{1}{1+\mathrm{NMAE}},
\label{eq:level}
\end{align}
where $r_t$ is the daily log return of the target in percent. The $\sqrt{T}$ term
reflects the horizon over which deviations accumulate, and normalizing by the
target's own volatility makes the score comparable across assets.

\paragraph{Shape.}
The shape score is the Spearman rank correlation between the realized and target paths,
\begin{equation}
\mathrm{Shape}
=
\frac{1}{2}\left(1+\rho_{\mathrm{Spearman}}(P,M)\right),
\label{eq:shape}
\end{equation}
which lies in $[0,1]$. Because Spearman correlation depends only on ranks, it is
invariant to monotone rescaling and rewards agreement in the ordering of the path,
so a path that is too volatile but correctly ordered is not penalized twice;
amplitude is handled by the level component.

\paragraph{Realization score (the tracking score).}
\begin{equation}
\Q=\frac{1}{2}\,\mathrm{Level}+\frac{1}{2}\,\mathrm{Shape}.
\label{eq:Q}
\end{equation}
Level and shape contribute equally, and both are bounded in $[0,1]$, so $\Q\in[0,1]$. The two components capture complementary properties: level measures amplitude accuracy and shape measures temporal ordering. The equal weighting is fixed before the experiments and is not tuned on the reported results.

\subsection{True and Flipped Targets}
\label{sec:evaluation:targets}

For each asset or scenario $i$, let $c_{i,t}$ be the close price on day $t$ and $c_{i,0}$ the close on the normalization day. The true normalized path is
\begin{equation}
\MT_{i,t} = \frac{c_{i,t}}{c_{i,0}}\cdot 1000,
\label{eq:true}
\end{equation}
and the flipped path is
\begin{equation}
\MF_{i,t} = 2000-\MT_{i,t}.
\label{eq:flip}
\end{equation}
Both paths start at $1000$ with the same absolute deviation at each date but
opposite directions; the construction creates two counterfactual announcements of identical form and
frequency, so the model cannot tell them apart, and the true/flipped label exists
only on the evaluator side.

\subsection{Two Observable Scores}
\label{sec:evaluation:scores}

Define the true-signal realization score and the flipped-signal realization score as
\begin{align}
\mathrm{TSR} &= \E_i\left[\Q\left(\W(\MT_i),\MT_i\right)\right],\label{eq:tsr}\\
\mathrm{FSR} &= \E_i\left[\Q\left(\W(\MF_i),\MF_i\right)\right].\label{eq:fsr}
\end{align}
Each arm is scored against its own target; scoring the flipped arm against the true target would mechanically lower FSR and is therefore invalid.

\subsection{Reference-Only Regime}
\label{sec:evaluation:zero}

In this regime no external news or fundamental information is injected. The true and flipped references are symmetric, so, ideally,
\begin{equation}
\mathrm{TSR}\approx\mathrm{FSR}.
\label{eq:symmetry}
\end{equation}
Their common level defines realization capability:
\begin{equation}
\R=\frac{\mathrm{TSR}+\mathrm{FSR}}{2},
\label{eq:R}
\end{equation}
Because the reference is present in both arms, $\R$ measures the ability to follow an announced macro target under the reference-only regime rather than the ability to realize a path from information alone. The paired gap defines structural bias,
\begin{equation}
\B=\sqrt{\E_i[\D_i^2]},\qquad
\D_i=\Q_i^{\mathrm{T}}-\Q_i^{\mathrm{F}}.
\label{eq:B}
\end{equation}
where $\Q_i^{\mathrm{T}}=\Q(\W(\MT_i),\MT_i)$ and $\Q_i^{\mathrm{F}}=\Q(\W(\MF_i),\MF_i)$.
A low $\B$ means that the configuration is close to neutral with respect to the direction of the path. A larger $\B$ means that the structure, constraints, or path construction introduces a directional bias.

$\B$ is a calibration quantity conditional on a given configuration, and it must be read
jointly with $\R$, because a model that realizes little in either arm can have a small $\B$
trivially.

\subsection{Information-Injection Regime}
\label{sec:evaluation:info}

In this regime external information is added. The information can alter micro behavior and can dilute the injected macro reference. Absolute TSR and FSR therefore no longer have a comparable realization interpretation. The interpretable quantity is the true--flipped gap
\begin{equation}
\D=\mathrm{TSR}-\mathrm{FSR}.
\label{eq:delta}
\end{equation}
A positive $\D$ means that the true announcement is more effective than its flipped mirror, and a larger $\D$ indicates a stronger information effect. The score is not perfectly invariant to path inversion, so a no-reference output can also have $\D\neq0$; we therefore read changes in $\D$ across conditions rather than the absolute value alone. In this regime, absolute TSR/FSR should not be used to judge the model, and $\B$ should not be treated as a structural-neutrality score.

\section{Experimental Design}
\label{sec:design}

\subsection{Testbed}
\label{sec:design:testbed}

The framework is domain-general; the empirical instantiation is a stock-market testbed built on the TwinMarket simulator \citep{yang2025twinmarket}. The evaluation suite contains five research composites of Chinese industry indices: consumer goods (CGEI), energy and resources (EREI), financial services (FSEI), real estate (REEI), and technology and telecom (TTEI). Each index is simulated independently with its own population of ten LLM agents and its own market, so the suite consists of five single-index simulations (50 agent instances in total); TSR and FSR average over the five market instances. The five indices were selected for dense news coverage; the platform's other five
sector indices have sparse coverage and are not evaluated. Each simulation covers 51 trading days (2023-06-15 to 2023-08-28) with five trading rounds per day. Agents are driven by DeepSeek-Chat under the platform's default sampling settings, and the initial stock/cash allocation is 70/30 unless stated otherwise. The reported cells average multiple independent runs per condition and asset.

Agents submit limit orders bounded by available cash and sellable shares, short
selling is disabled, and match prices are bounded by the $\pm10\%$ daily limit
around the previous close. In the continuous-trading rounds, decisions enter in a
randomly drawn intraday order, so orders are matched under price--time priority
rather than by simultaneous clearing; residual orders are cancelled before the
closing call auction, which starts from an empty book
(Appendix~\ref{app:implementation:market}). Micro behavior is implemented entirely
by the model.

The public reference is injected through the daily prompt: at the start of each trading day, agents see that day's normalized reference level. Agents do not observe future reference values, and whether and how they use the announced value is left to the model. The no-reference arm uses the same prompt without the reference block.

\subsection{Experimental Conditions}
\label{sec:design:conditions}

Each condition is run with both a true and a flipped reference (Table~\ref{tab:conditions}).
The 3:7, 70/30, news-off, fundamental-data-off cell is a shared reference-only
baseline; the news experiment uses the corresponding fundamental-data-on cell
(Appendix~\ref{app:runs}).

\begin{table}[t]
\centering
\footnotesize
\caption{Experimental conditions. Every condition is run on the five indices under both a true and a flipped macro reference. C1 varies the stock/cash split (structure), C2 the technical/fundamental composition (realization), and C3 whether fundamental data is injected (information).}
\label{tab:conditions}
\begin{tabular}{@{}p{0.17\columnwidth}p{0.42\columnwidth}p{0.29\columnwidth}@{}}
\toprule
Family & Manipulation & Held fixed \\
\midrule
Control & true target with vs.\ without a public reference; news off vs.\ news on & composition 3:7, stock/cash 70/30, fundamental data on \\
C1 & stock/cash $=$ 30/70, 50/50, 70/30, 90/10 & composition 3:7, news off, fundamental off \\
C2 & technical/fundamental $=$ 0/10, 3/7, 5/5, 10/0 & stock/cash 70/30, news off, fundamental off \\
C3 & fundamental information off vs.\ on & composition 3:7, stock/cash 70/30, news off \\
News & news off vs.\ on & composition 3:7, stock/cash 70/30 \\
\bottomrule
\end{tabular}
\end{table}

\subsection{Protocol}
\label{sec:design:protocol}

Every path is evaluated on the 51 simulation days only: including pre-simulation
history would make the true target coincide with the observed history and favor the
true arm. TSR and FSR are the means of $\Q$ over the five assets, $\B$ is the root-mean-square of the paired differences, and $\D$ is the difference of the two means.

The no-reference baseline uses the mean no-reference output per asset, evaluated
against both targets. Each reported condition is averaged over multiple
independent stochastic runs per asset and arm, and we report asset-level bootstrap
intervals for the aggregate quantities (Section~\ref{sec:results:validation}).

\section{Results}
\label{sec:results}

\subsection{Control Effectiveness}
\label{sec:results:control}

\begin{figure}[t]
  \centering
  \includegraphics[width=\columnwidth]{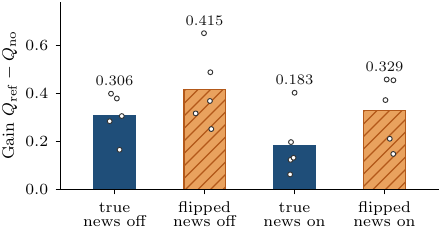}
  \caption{Per-asset control gains for the four groups (true and flipped reference, news off and on). Bars are group means and dots the five individual assets; solid bars are true references and hatched bars flipped references. Every per-asset gain is positive, and the group means are $0.306$ and $0.415$ without news and $0.183$ and $0.329$ with news.}
  \label{fig:control}
\end{figure}

Table~\ref{tab:control} and Figure~\ref{fig:control} report control effectiveness for the
four groups (true and flipped reference, news off and on); Appendix~\ref{app:per-asset} gives the
per-asset tables. The announcement shifts the output toward the announced path in every asset and group; since the no-reference arm is at the zero-information level on the true target
(Section~\ref{sec:results:validation}), this gain reflects the announcement rather than
a capability of the model. Without news, the gain is $0.306$ on the true reference (per-asset range $0.165$--$0.399$, standard deviation $0.093$) and $0.415$ on the flipped reference (range $0.251$--$0.651$, standard deviation $0.158$). With news, the true-reference gain is $0.183$ (range $0.062$--$0.403$, standard deviation $0.132$) and the flipped-reference gain is $0.329$ (range $0.148$--$0.458$, standard deviation $0.142$).

Both news-on gains are smaller than their news-off counterparts, which does not indicate a
failure of control:
injecting news dilutes the announced reference, which must compete with other information
for its effect on the output, and the reference-arm score falls in both arms ($0.9180\to0.8307$ true, $0.8839\to0.7820$
flipped). A smaller gain therefore
reflects a weaker marginal effect of the announcement, not an inability to steer the market.

Both announced references are followed, and the flipped-reference gain is larger than the true-reference gain. The difference is mechanical: the gain is bounded by $1-\Q_{\mathrm{no}}$, and the flipped arm starts from a lower no-reference score ($0.4688$ versus $0.6115$). This is consistent with controllability, although the model may also be echoing the
announced path; it is why $\R$ is read as realization \emph{under the announced
reference}.

\begin{table}[t]
\centering
\small
\caption{Control effectiveness for the four groups (true and flipped reference
$\times$ news off and on). Both news conditions include the fundamental data and
therefore lie in the information-injection regime; the reference-only baseline of
Tables~\ref{tab:c1}--\ref{tab:c2} is a different cell. $\Q_{\mathrm{no}}$ is the
no-reference score on the announced reference, $\Q_{\mathrm{ref}}$ the score with
the public reference, and Gain their difference (from unrounded values).}
\label{tab:control}
\begin{tabular}{@{}llccc@{}}
\toprule
Condition & Reference & $\Q_{\mathrm{no}}$ & $\Q_{\mathrm{ref}}$ & Gain \\
\midrule
News off & true & 0.6115 & 0.9180 & 0.3065 \\
         & flipped & 0.4688 & 0.8839 & 0.4152 \\
\addlinespace
News on  & true & 0.6473 & 0.8307 & 0.1834 \\
         & flipped & 0.4533 & 0.7820 & 0.3287 \\
\bottomrule
\end{tabular}
\end{table}

\subsection{Structural Bias}
\label{sec:results:structure}

Table~\ref{tab:c1} reports the reference-only stock/cash experiment (Figure~\ref{fig:scan}(a), Appendix~\ref{app:figs}). Structural bias is non-monotonic. It falls from $0.2277$ at 30/70 to $0.0418$ at 70/30, then rises to $0.1831$ at 90/10. The 70/30 allocation is the lowest-bias point in this scan (the composition scan reaches a lower bias at 0/10,
$\B_{\mathrm{rms}}=0.0169$, with slightly lower realization), and it is also the point where TSR and FSR are nearly equal ($0.8917$ and $0.8970$). At 30/70 and 90/10 the TSR/FSR gaps widen, and the sign of the mean gap changes
across the scan, from positive at low stock exposure to negative at high stock
exposure---consistent with the long-only and inventory constraints rather than
with a fixed property of the model.

\begin{table}[t]
\centering
\small
\caption{Reference-only stock/cash experiment. TSR and FSR are means over the five assets; $\R$ is their average; $\bar{\D}$ is the mean paired gap and $\B_{\mathrm{rms}}$ is its root mean square; unrounded values.}
\label{tab:c1}
\setlength{\tabcolsep}{4pt}%
\begin{tabular}{@{}lccccc@{}}
\toprule
Stock/cash & TSR & FSR & $\R$ & $\bar{\D}$ & $\B_{\mathrm{rms}}$ \\
\midrule
30/70 & 0.7802 & 0.7258 & 0.7530 & $+0.0544$ & 0.2277 \\
50/50 & 0.8527 & 0.8222 & 0.8375 & $+0.0305$ & 0.1184 \\
70/30 & 0.8917 & 0.8970 & 0.8943 & $-0.0052$ & 0.0418 \\
90/10 & 0.6890 & 0.7884 & 0.7387 & $-0.0994$ & 0.1831 \\
\bottomrule
\end{tabular}
\end{table}

\subsection{Realization Capability}
\label{sec:results:realization}

Table~\ref{tab:c2} reports the reference-only composition experiment (Figure~\ref{fig:scan}(b), Appendix~\ref{app:figs}). $\R$ is highest at 3:7 ($0.8943$), slightly above 0:10 ($0.8873$). It falls to $0.8134$ at 5:5 and collapses to $0.3581$ at 10:0. A small technical component raises the score, while technical dominance causes a clear failure; within the scanned range the relation is non-monotone and peaks at 3:7.

The 10:0 condition is especially informative: its TSR is $0.2764$ and its FSR
$0.4398$, so the mean paired gap is $-0.1633$. A technical-only structure cannot
realize the reference path and the true arm is weaker than the flipped one, so its
structural bias comes from technical failure rather than from genuine
information---consistent with reading $\B$ jointly with $\R$
(Section~\ref{sec:evaluation:zero}).

Read across conditions, this is the central realizability result: the same
announced macro target is reached at 3:7 ($\R=0.8943$) and not at 10:0
($\R=0.3581$), so realizing an announced target is not automatic---it depends on
the micro structure of the market.

\begin{table}[t]
\centering
\small
\caption{Reference-only composition experiment. $\R=(\mathrm{TSR}+\mathrm{FSR})/2$; $\bar{\D}$ and $\B_{\mathrm{rms}}$ are the mean and root-mean-square paired gaps; unrounded values.}
\label{tab:c2}
\setlength{\tabcolsep}{4pt}%
\begin{tabular}{@{}lccccc@{}}
\toprule
Tech./fund. & TSR & FSR & $\R$ & $\bar{\D}$ & $\B_{\mathrm{rms}}$ \\
\midrule
0/10 & 0.8829 & 0.8917 & 0.8873 & $-0.0089$ & 0.0169 \\
3/7  & 0.8917 & 0.8970 & 0.8943 & $-0.0052$ & 0.0418 \\
5/5  & 0.8359 & 0.7908 & 0.8134 & $+0.0451$ & 0.0871 \\
10/0 & 0.2764 & 0.4398 & 0.3581 & $-0.1633$ & 0.2347 \\
\bottomrule
\end{tabular}
\end{table}

\subsection{Information Effectiveness}
\label{sec:results:info}

Table~\ref{tab:c3} reports the fundamental-information experiment. The first row is reference-only: the two arms are symmetric and the mean paired gap is $\bar{\D}=-0.0052$, close to zero, as the symmetry assumption predicts. The second row adds fundamental data and therefore lies in the information-injection regime; there the interpretable quantity is $\D=+0.0341>0$, indicating that the injected information is used directionally. Absolute TSR and FSR also change, but under information injection they are
diluted and are not used as realization measures. Because the two rows belong to
different regimes (Table~\ref{tab:regimes}), we do not interpret their
difference; the effect is read from changes in $\D$ across injection-regime conditions,
including the no-reference arm (Section~\ref{sec:results:news}).

\subsection{News as an Information Channel}
\label{sec:results:news}

\begin{figure}[t]
  \centering
  \includegraphics[width=\columnwidth]{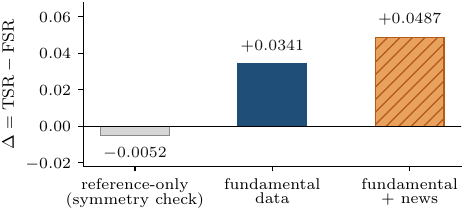}
  \caption{Information effects measured by $\D=\mathrm{TSR}-\mathrm{FSR}$. The first bar is the reference-only symmetry check; the second adds fundamental data and the third adds news to the injected information. Both information-injection bars are positive, and news raises $\D$ from $0.0341$ to $0.0487$. Absolute scores are not interpreted once information is injected.}
  \label{fig:channels}
\end{figure}

Table~\ref{tab:news} and Figure~\ref{fig:channels} report the news experiment. Both conditions include the fundamental data, so both lie in the information-injection regime and their $\D$ values are comparable. Adding news to the injected information raises $\D_{\mathrm{ref}}$ from $0.0341$ to $0.0487$, i.e., it increases the true-over-flipped advantage and thus the information effect measured in this regime.

The last three columns of Table~\ref{tab:news} report an auxiliary check outside
the framework: the same no-reference output scored against both targets. On this
baseline $\Q_{\mathrm{T}}$ rises from $0.6115$ to $0.6473$ and $\Q_{\mathrm{F}}$
falls from $0.4688$ to $0.4533$, so $\D_{\mathrm{no}}$ rises from $0.1427$ to
$0.1940$, in the same direction as the reference-condition result. Because
$\D_{\mathrm{no}}$ is not zero, $\D$ is not a pure information measure, so the
identification rests on the within-condition comparison (news off versus on),
which moves in the same direction in both arms.

The absolute reference-condition TSR/FSR fall, but this is not evidence against
news: once information is injected the reference is diluted, so the effect is
carried by $\D$. The no-reference arm moves in the same direction, which is
consistent with news acting as a corrective signal rather than as a replacement for
the reference.

\subsection{Calibration and Robustness}
\label{sec:results:validation}

We calibrate the scores against trivial paths, alternative score definitions,
repeated batches and bootstrap intervals (Appendix~\ref{app:validation}). A constant path scores $0.610$, a straight line
between endpoints $0.821$ and a volatility-matched random walk $0.565$, all below the reference arm in the news-off cell ($0.9180$). The no-reference arm itself lies at the
constant-path and random-walk level ($0.6115$ versus $0.6104$, 58th percentile of
the matched-volatility null), with a better ordering but a worse level: without an
announced reference the absolute score carries no realization information, which
is exactly the confusion that the two-regime framework prevents. Under five alternative definitions of $\Q$, $\R$ stays in $0.760$--$0.924$ and $\B$ in $0.036$--$0.051$.

\subsection{Summary of Findings}
\label{sec:results:summary}

The experiments support four conclusions. \textbf{(F1, control)} An announced public
reference shifts the output toward the announced path in every asset and
information condition. \textbf{(F2, structure)} Within the stock/cash scan, a 70/30 allocation gives the lowest structural
bias and near-equal TSR/FSR, with bias non-monotonic in the allocation. \textbf{(F3, realization)}
Realization is non-monotone in the technical share and peaks at 3:7 within the
scanned range. \textbf{(F4, information)} In the injection regime $\D$ is $0.0341$ with fundamental data
and $0.0487$ with news, and the no-reference baseline agrees in direction.

\section{Discussion and Limitations}
\label{sec:discussion}

\paragraph{Control and capability are different questions.}
The control experiment asks how much the announcement moves the output toward a target;
the capability measurements ask how the two stages behave when the announcement is
present in both arms. A model may follow the announced path and still be structurally
biased, or realize it without using genuine information.

\paragraph{The same scores need different interpretations.}
TSR and FSR are observable in every experiment, but their meaning depends on the
regime: reference-only scores estimate realization and structural bias; after
information injection only their difference is interpretable. We state the regime
with every table.

\paragraph{Scope and limitations.}
The framework is domain-general, but the evidence comes from one simulator, one
LLM, one 51-day window and five narrow composites, one of which is a
single-stock basket; the reported means average multiple independent runs, and the
reference is an evaluator-side target rather than a forecast. The operating point is a region in configuration space, not a universal recommendation; seed-level
variance, lagged references and other LLM backbones are natural next steps.

\section{Conclusion}
\label{sec:conclusion}

This paper studied macro control and capability evaluation for black-box world
models. A public macro reference provides soft control without touching micro
decisions, and a two-regime framework uses the same observable scores for
different purposes: reference-only TSR/FSR give realization under the announced
reference and structural bias, while information-injection TSR/FSR give the
differential signal $\D$. On five single-index LLM-agent markets, an announced
public reference shifts the output toward the announced path in every asset and
condition, while the no-reference score on the true target remains near the null level ($0.61$--$0.65$); the score is non-monotone in the technical-agent share and peaks at 3:7; and $\D$ is $0.0341$ with fundamental
information and $0.0487$ once news is added. Control and capability are distinct properties, and absolute path scores
become uninterpretable once external information is injected.

\section*{Impact Statement}

The framework is intended to make LLM-agent world models easier to audit from their macro
behavior alone: a public reference path gives a transparent control interface, so an
evaluator can test whether a model follows an announced target without access to its internal
decisions. The procedure uses only macro inputs and outputs and can be applied to proprietary
models without inspecting their agent-level implementation. All simulations are synthetic and
use public historical data; nothing here is investment advice or a recommendation to deploy the system in real markets. We see no
direct harmful
application of the protocol itself; the broader societal impact depends on the models and
information channels to which it is applied.

\bibliographystyle{icml2026}
\bibliography{refs_rev2}

\clearpage
\appendix
\IfFileExists{appendix_rev2.tex}{\input{appendix_rev2.tex}}{%
  \section{Appendix placeholder}
  To be generated.
}
\input{appendix_extra}

\end{document}

%% file: figs/fig1_framework_tikz.tex
\begin{tikzpicture}[
  x=1pt, y=1pt,
  font=\footnotesize,
  >={Stealth[length=1.5mm,width=1.1mm]},
  box/.style={draw, rounded corners=1pt, align=center, inner sep=2.2pt, line width=0.5pt},
  wbox/.style={box, fill=black!10},
  dashbox/.style={draw, dashed, rounded corners=1pt, align=center, inner sep=2.2pt, line width=0.5pt},
  arm/.style={->, line width=0.6pt},
  darm/.style={->, dashed, line width=0.5pt},
  lbl/.style={font=\scriptsize, align=center, inner sep=0.6pt},
  pttl/.style={anchor=north west, font=\footnotesize\bfseries},
  frame/.style={rounded corners=1.5pt, line width=0.5pt}
]

\node[box] (in) at (-79,0) {macro input\\ $M_{\mathrm{in}}$};
\node[wbox, minimum width=92pt] (w) at (0,0)
  {micro behavior\\ $B_{\mathrm{micro}}$ (internal)\\[0.6pt] \scriptsize unobserved by evaluator};
\node[box] (out) at (81,0) {macro output\\ $M_{\mathrm{out}}$};
\draw[arm] (in) -- (w);
\draw[arm] (w) -- (out);
\node[lbl, above=0.5pt of w] {black-box world model $\W$};
\node[lbl, below=3pt of w] (ref)
  {public macro reference $M^{\mathrm{ref}}$ (announced each trading day)};
\draw[darm] (ref.north) -- (w.south);

\draw[frame] (-114,-34) rectangle (54,-134);
\node[pttl] at (-111,-36) {(a) reference-only regime};

\node[box, minimum width=60pt] (t1) at (-72,-59) {true ref\\ $M^{\mathrm{T}}$};
\node[wbox, minimum width=13pt, minimum height=10pt] (w1) at (-19,-59) {$\W$};
\node[anchor=west] (tsr) at (-7,-59) {$\mathrm{TSR}$};
\draw[arm] (t1.east) -- (w1.west);
\draw[arm] (w1.east) -- (tsr.west);

\node[box, minimum width=60pt] (t2) at (-72,-83) {flipped ref\\ $M^{\mathrm{F}}=2000-M^{\mathrm{T}}$};
\node[wbox, minimum width=13pt, minimum height=10pt] (w2) at (-19,-83) {$\W$};
\node[anchor=west] (fsr) at (-7,-83) {$\mathrm{FSR}$};
\draw[arm] (t2.east) -- (w2.west);
\draw[arm] (w2.east) -- (fsr.west);

\node[anchor=west, align=left] at (-106,-103)
  {$R=\tfrac{1}{2}(\mathrm{TSR}+\mathrm{FSR})$\ \ (common level)};
\node[anchor=west, align=left] at (-106,-120)
  {$B=\sqrt{\E_i[\D_i^2]}$\ \ (RMS paired gap)};

\draw[frame] (-114,-140) rectangle (80,-234);
\node[pttl] at (-111,-146) {(b) information-injection regime};

\node[box, minimum width=84pt] (r1) at (-64,-170)
  {reference path $M^{\mathrm{ref}}$\\ (true\,/\,flipped)};
\node[wbox, minimum width=13pt, minimum height=10pt] (w3) at (2,-170) {$\W$};
\node[anchor=west, align=left] (tf) at (14,-170) {$\mathrm{TSR}$, $\mathrm{FSR}$};
\draw[arm] (r1.east) -- (w3.west);
\draw[arm] (w3.east) -- (tf.west);

\node[dashbox, minimum width=88pt] (ext) at (2,-198)
  {external information\\ news\,/\,fundamental data};
\draw[darm] (ext.north) -- (w3.south);

\node[anchor=west, align=left] at (-108,-216)
  {interpret only $\D=\mathrm{TSR}-\mathrm{FSR}$};
\node[anchor=west, align=left] at (-108,-227)
  {absolute $\mathrm{TSR}$, $\mathrm{FSR}$ not comparable};

\end{tikzpicture}

%% file: appendix_rev2.tex

\section{Evaluation Details}
\label{app:audit}

\subsection{Evaluation Window and Arm-Specific Targets}
\label{app:audit:window}

Every metric is computed on the 51 trading days of the simulation window,
2023-06-15 to 2023-08-28, and on nothing else. The data files start on
2023-01-03 and are used before the window only to build the targets, to
pre-compute the technical indicators and valuation series that agents may
query, and to normalize the price axis. No price observation from before the
window enters a score. Including warm-up history would make the true target
coincide with the observed history and would mechanically favor the true arm.

For each asset $i$, let $c_{i,t}$ be the index close on day $t$. The true
target path is
\begin{equation}
\MT_{i,t} = \frac{c_{i,t}}{c_{i,0}}\cdot 1000,
\label{eq:app-true}
\end{equation}
where $c_{i,0}$ is the close on the \emph{normalization day}
2023-06-14, the last trading day before the simulation window. The
flipped target is $\MF_{i,t}=2000-\MT_{i,t}$. The normalization day is chosen
so that the target level is exactly 1000 on the day before trading starts; the
simulated price series is normalized with the same base, which makes the
simulated and target paths share one price axis. In particular, the targets do
\emph{not} start at 1000 on the first scored day: the first scored level equals
1000 plus the real first-day return relative to the base, e.g.\ $\MT_{\mathrm{CGEI}}=
1016.58$ and $\MF_{\mathrm{CGEI}}=983.42$ on 2023-06-15. Both targets coincide
at 1000 only on the normalization day, which is outside the scored window.

Each arm is scored against its own target: the true arm against $\MT$ and the
flipped arm against $\MF$. Scoring the flipped arm against $\MT$ would produce
a mechanically low FSR and is therefore invalid; the same rule is used for the
flipped-arm control gain in Appendix~\ref{app:per-asset}. TSR and FSR are
means of $\Q$ over the five independent single-asset runs, $\B$ is the
root-mean-square of the paired true--flipped gaps, and $\D$ is the difference
of the two means. All quantities are descriptive: each cell averages multiple independent
runs per asset and arm (Section~\ref{sec:discussion}).

\subsection{Level Normalization}
\label{app:audit:level}

Let $P$ be a realized path and $M$ its target path over $T=51$ days. The daily
relative error in percent and the target log-return volatility are
\begin{align}
e_t&=\left|\frac{P_t}{M_t}-1\right|\times100,\\
r_t&=100\,(\log M_t-\log M_{t-1}),\qquad
\sigma_M=\operatorname{std}(r_t),
\label{eq:app-err}
\end{align}
and the level component is
\begin{equation}
\mathrm{NMAE}=\frac{\operatorname{mean}(e_t)}{\sigma_M\sqrt{T}},\qquad
\mathrm{Level}=\frac{1}{1+\mathrm{NMAE}}.
\label{eq:app-level}
\end{equation}
Two design choices are worth stating explicitly. First, the $\sqrt{T}$ term
matches the horizon over which independent daily errors accumulate: with a
daily denominator the level score is compressed toward zero as $T$ grows and
becomes incomparable across evaluation windows; with $\sigma_M\sqrt{T}$ it is
on a comparable scale for $T=51$. Second, the denominator uses the target's
own volatility, which makes assets with different volatility comparable: over
the five targets $\sigma_M$ ranges from $0.93\%$/day (EREI) to $1.98\%$/day
(REEI) and the flipped target is not identical in volatility to the true target
(the flipped/true ratio is $1.01$--$1.06$ for four assets and $0.89$ for TTEI,
which fell over the window). Because the two arms are scored against different
paths, each arm's level is normalized by its own target volatility. A variant
that uses one common volatility for both arms is reported in
Appendix~\ref{app:validation}.

\subsection{Shape}
\label{app:audit:shape}

The shape component is the Spearman rank correlation of the realized and target
paths,
\begin{equation}
\mathrm{Shape}=\frac{1}{2}\bigl(1+\rho_{\mathrm{Spearman}}(P,M)\bigr)\in[0,1].
\label{eq:app-shape}
\end{equation}
Spearman correlation uses ranks only, so the common normalization level (1000)
and any monotone rescaling cancel; it rewards agreement in the ordering of the
path rather than its amplitude, which is measured separately by Level. We do
not use a level-based Pearson correlation, which would be dominated by the
shared trend and would be insensitive to the ordering of day-to-day changes. A
variant that computes Shape on daily returns instead of levels is reported in
Appendix~\ref{app:validation}.

\subsection{Equal Weighting and Component Decomposition}
\label{app:audit:weighting}

The realization score is $\Q=\tfrac12\mathrm{Level}+\tfrac12\mathrm{Shape}$.
Both components are bounded in $[0,1]$, so $\Q\in[0,1]$ and the equal weighting
was fixed before the experiments and never tuned on the reported results.
Table~\ref{tab:app-components} decomposes the headline reference-only and
information-injection cells into their two components. Three observations
follow. (i) Level and Shape move in the same direction in these cells: the true
arm has both a higher level and a higher shape than the no-reference baseline,
so the conclusion does not depend on one component dominating. (ii) The
flipped arm's shape drops sharply once news is injected ($0.9235\to0.7777$),
whereas the true arm's shape falls less ($0.945\to0.861$); this is the main
reason the true--flipped gap rises under news. (iii) For the true arm the two
components fall by similar amounts when news is injected
(Level $0.891\to0.801$, Shape $0.945\to0.861$), so the 50/50 weighting is not
driving the aggregate movement.

\begin{table}[t]
\centering
\footnotesize
\caption{Component decomposition of the headline cells (means over the five
assets). The true arm is scored against $\MT$ and the flipped arm against
$\MF$; TSR and FSR are the corresponding means of $\Q$, equal to the average
of the two components up to rounding; the reference-only column has fundamental data off,
and both news columns have it on.}
\label{tab:app-components}
\setlength{\tabcolsep}{4pt}%
\begin{tabular}{@{}lccc@{}}
\toprule
Component & Ref.-only & News off & News on \\
\midrule
Level, true arm & 0.8645 & 0.8906 & 0.8008 \\
Shape, true arm & 0.9189 & 0.9453 & 0.8605 \\
TSR & 0.8917 & 0.9180 & 0.8307 \\
\midrule
Level, flipped arm & 0.8658 & 0.8443 & 0.7863 \\
Shape, flipped arm & 0.9281 & 0.9235 & 0.7777 \\
FSR & 0.8970 & 0.8839 & 0.7820 \\
\bottomrule
\end{tabular}
\end{table}

\subsection{Notation}
\label{app:notation}

Table~\ref{tab:notation} collects the notation used in the remainder of the
paper.

\begin{table}[t]
\centering
\footnotesize
\caption{Notation.}
\label{tab:notation}
\setlength{\tabcolsep}{4pt}%
\begin{tabular}{@{}l p{0.60\columnwidth}@{}}
\toprule
Symbol & Meaning \\
\midrule
$\Min$, $\Bm$, $\Mout$ & macro information, micro behavior, macro output \\
$\W$, $P=\W(M)$ & black-box world model and its macro output path \\
$\Mref$ & public macro reference path announced by the evaluator \\
$\MT$, $\MF$ & true target path and flipped target path \\
$\Q(P,M)$ & path realization score of output $P$ against target $M$ \\
$e_t$, $\sigma_M$ & daily relative error (\%); volatility of target log returns (\% per day) \\
$\mathrm{Level}$, $\mathrm{Shape}$ & level and shape components of $\Q$ \\
$\mathrm{TSR}$, $\mathrm{FSR}$ & mean realization under true / flipped targets \\
$\R$ & realization under the announced reference, $(\mathrm{TSR}+\mathrm{FSR})/2$ \\
$\D$ & true--flipped gap, $\mathrm{TSR}-\mathrm{FSR}$ \\
$\B$ & structural bias, root-mean-square of paired gaps \\
$\E_i[\cdot]$ & average over the evaluated assets \\
\bottomrule
\end{tabular}
\end{table}

\subsection{Interpretation Rules}
\label{app:rules}

\begin{table}[t]
\centering
\small
\caption{Interpretation rules for the same two observable scores.}
\label{tab:regimes}
\begin{tabular}{@{}p{0.25\columnwidth}p{0.27\columnwidth}p{0.31\columnwidth}@{}}
\toprule
Regime & Quantities & Interpretation \\
\midrule
Reference-only & TSR, FSR, $\R$, $\B$ & Common level measures realization capability; paired gap measures structural bias; $\B$ is read jointly with $\R$. \\
Information-injection & $\D=\mathrm{TSR}-\mathrm{FSR}$ & Positive or increasing $\D$ indicates effective information; absolute TSR/FSR are not interpreted. \\
\bottomrule
\end{tabular}
\end{table}

\section{Implementation Details}
\label{app:implementation}

\subsection{Platform, Runs, and Agents}
\label{app:implementation:platform}

The testbed builds on the open-source TwinMarket simulator
\citep{yang2025twinmarket}. A run instantiates one tradable index market with
its own order book, its own price and news data, and ten LLM agents. The ten
agents are a fixed-seed subset ($\text{seed}=0$) of the platform's synthetic
retail-investor population; each has a persona (demographics, risk attitude,
investment style) and an account holding the index and cash. In the reported
conditions the initial allocation is $70\%$ stock and $30\%$ cash at the
normalized price of 1000 points (7{,}000 shares and CNY~3M cash for the initial
CNY~10M account); all conditions share the same initial states. Each evaluated
index is an independent run, and the five runs of a condition differ only in
the index series, its news stream and the agent subset generated for that
asset. Data preparation reads the archived ten-sector price file for the asset
(484 trading days, 2023-01-03 to 2024-12-31), draws the ten-user subset from the
1{,}000-user seed database with seed~$0$, and rescales the series so that the
2023-06-14 close equals 1{,}000 points; for CGEI the scaling factor is $103.627$
(raw base close $9.65$). The initial account is CNY~10M, $70\%$ invested, i.e.\
7{,}000 index shares at 1{,}000 points and CNY~3M cash.

Two agent attributes are manipulated. The stock/cash ratio sets the initial
allocation. The technical/fundamental composition sets the persona and analysis
mode of the ten agents: in a $3{:}7$ configuration, three agents have a
technical-trader persona and seven a fundamental-investor persona. Personas
also differ in risk attitude and trading style and are held fixed across
conditions. Every reported run uses the full LLM pipeline for all ten agents;
the rule-based ``random trader'' fallback present in earlier versions of the
platform is disabled in these runs.

\subsection{Daily Agent Workflow}
\label{app:implementation:workflow}

A simulation day has five trading rounds. Before the first round, each agent
runs through the following stages; only stages enabled in the condition are
executed.

\begin{enumerate}[leftmargin=*,itemsep=1pt,topsep=2pt]
  \item \textbf{Persona and account snapshot.} The system prompt fixes the
  persona; the first user message contains the account state (total assets,
  cash, position, holding cost, cumulative P\&L) and the current market
  snapshot.
  \item \textbf{News analysis} (news condition only). The day's public news
  list is delivered to every agent because the social graph is disabled and
  therefore all agents count as top users. The agent writes a short
  interpretation of the news.
  \item \textbf{News retrieval} (news condition only). The agent formulates a
  keyword query, the system retrieves the most relevant items from the indexed
  news database over the previous seven days, and the agent summarizes the
  retrieved items.
  \item \textbf{Indicator summary} (indicator conditions). The agent selects
  indicators from the available list; the system returns a daily series ending
  at the previous close, and the agent writes an objective summary.
  \item \textbf{Analysis.} The agent analyses the index given the information
  collected above. In the reference arm this is where the day's public
  reference value is announced (Appendix~\ref{app:implementation:prompt}).
  \item \textbf{Five decision rounds.} In each round the agent sees the current
  account and order-book state, the stage rules, the public reference, and the
  asset details; it returns at most one order, either a buy or a sell, with a
  position size in percent of total assets and a worst acceptable price.
\end{enumerate}

Belief updates, forum reading/interaction, posting and stock selection are
disabled in the reported conditions, so they add no model calls and no
contamination across agents. Because there is one tradable asset per run, the
selection stage is also without effect.

\subsection{Model, Sampling, and Call Volume}
\label{app:implementation:model}

Agents are driven by DeepSeek-Chat through the provider's OpenAI-compatible
chat API. Sampling uses the platform's call defaults: temperature $0.9$,
top-$p$ $0.9$, frequency penalty $0.5$, presence penalty $0.5$, and a maximum of
4096 generated tokens. The indicator-summary call is the only active stage that
overrides the default, with temperature $0.2$ because its output is an objective
data summary. No sampling seed is passed, so every run is stochastic. The model
string is \texttt{deepseek-chat} and the base URL is
\url{https://api.deepseek.com}.

Two reproducibility limitations attach to the model. First, \texttt{deepseek-chat}
is a provider-side rolling alias: no specific model snapshot was exposed in the
API response or recorded in the run artifacts, so an exact rerun at a later date
is not guaranteed to use the same weights. Second, because the model was called
through the provider's hosted service, the backend version behind the alias is
outside our control; the archived prompts and responses are released so that the
exact inputs and outputs of every call can be inspected.

Call volume is predictable from the pipeline. A news-off run makes
$10\times51\times(1+1+5)=3{,}570$ calls (indicator summary, analysis, and five
decisions per agent-day), and a news-on run adds four calls per agent-day (one
news analysis and three retrieval steps), for $10\times51\times11=5{,}610$ calls.

\subsection{Market Rules}
\label{app:implementation:market}

The five rounds of a day are an opening call auction, three continuous-trading
rounds, and a closing call auction. Their mechanics are those of a
Chinese-A-share-style limit-order market.

\paragraph{Orders.} In each round an agent may submit at most one order, a buy
or a sell with a position size in percent of total assets and a worst
acceptable price. The worst price is a marketable limit: a buy is executed
against the best available sell levels up to its worst price, a sell against
the best available buy levels down to its worst price, and execution proceeds
level by level with price impact. In the two call auctions all valid orders are
aggregated into a single clearing price. In the three continuous-trading rounds,
each decision is assigned a random timestamp within the trading sessions
(09:30--11:30 and 13:00--15:00); timestamps are made unique and sorted, and the
resulting order stream is applied to the book one order at a time under
price--time priority. Random timestamp assignment makes the execution order
within a round stochastic rather than simultaneous. Position sizes are converted into
whole 100-share lots (the A-share minimum) and capped by available cash on the
buy side and by sellable shares on the sell side.

\paragraph{Price limits.} Order prices and match prices are clipped to the
$\pm10\%$ daily limit around the previous close. Orders outside this band are
not generated or are clipped into it, and match prices are bounded by the same
band.

\paragraph{Long-only and T+1.} Short selling is disabled. T+1 settlement is
enforced in the order budget: shares bought and filled today cannot be sold
today (the sellable quantity is the day-start holding minus shares already sold
today), and cash proceeds from a sale are not available for a same-day purchase
(the buy budget is day-start cash minus current-day purchases). Same-day sale
proceeds therefore cannot be recycled into a new position on the same day.

\paragraph{Residual orders.} Unfilled parts of an order can remain in the book
between consecutive rounds; the agent may cancel them in a later round using
the order identifiers it was given. The last continuous round auto-cancels all
residual orders before the closing call auction, which therefore starts from
an empty book, and any order still unfilled at the end of the day expires.
The main text summarizes this as automatic cancellation of residual orders.

\subsection{Data}
\label{app:implementation:data}

\paragraph{Index series.} Each of the five evaluated indices is a
fixed-weight basket of representative A-share constituents. The composite daily
series were constructed by the authors from constituent prices; they are
research composites rather than official exchange indices, and the five
evaluated assets are a news-dense subset of the platform's ten sector
composites (the other five, TLEI, MEI, CPEI, IEEI and TSEI, have sparse news
coverage and are not evaluated). Table~\ref{tab:app-assets} lists the five
assets and their largest constituent. The data file provides, per index and
trading day, the close, previous close, percentage change, turnover, net
ultra-large order flow, PE/PB/PS/dividend-yield valuations, and 5/10/30-day
moving averages and turnover averages; it covers 2023-01-03 to 2024-12-31 (484
trading days per index), of which the 51-day evaluation window is used for
scoring.

\begin{table*}[t]
\centering
\small
\caption{The five evaluated composite indices. ``Constituents'' is the number
of A-share stocks in the fixed-weight basket; the parenthesised weight is the
largest constituent's share within the index. The last column is the sector's
weight in the platform's aggregate ten-sector total-market index; it is not
used in the single-index runs and is listed only for provenance.}
\label{tab:app-assets}
\setlength{\tabcolsep}{6pt}%
\begin{tabular}{@{}llllc@{}}
\toprule
Code & Name & Example constituents & Const. & Platform \\
\midrule
CGEI & consumer goods & Kweichow Moutai (69.2\%) & 5 & 18.2\% \\
EREI & energy \& resources & China Yangtze Power (33.5\%) & 6 & 13.2\% \\
FSEI & financial services & Ping An (22.9\%) & 11 & 33.0\% \\
REEI & real estate & Poly Developments (100\%) & 1 & 1.0\% \\
TTEI & technology \& telecom & SMIC (18.1\%) & 10 & 13.0\% \\
\bottomrule
\end{tabular}
\end{table*}

\paragraph{News stream.} The news condition uses a market-wide stream of
Chinese financial news. The stream contains 3{,}441 items over 550 calendar
days (2023-06-15 to 2024-12-15); the 75 calendar days covered by the evaluation
window contain 444 items (median 3 per calendar day, mean 5.9). News is
market-wide, not per-index: the local dataset does not provide per-industry
news, so all five single-index runs receive the same stream on the same
calendar day. When the news condition is enabled, each day's list is both
delivered as a public broadcast and indexed for the agents' own retrieval
queries. The retrieval index contains roughly $2.5\times10^5$ documents
(news items and exchange announcements), and an agent's query returns the most
relevant items over the preceding seven days. Two timing caveats apply. The
news store is timestamped at calendar-day granularity, so a day's list can
contain items published after the previous close; and the retrieval window ends
on the current date. Both give the news arm same-day information, a look-ahead
channel that is separate from (and weaker than) the announced reference
and that we did not control. A placebo (shuffled-news) arm was not run.

\paragraph{Fundamental and technical indicators.} When the indicator stage is
enabled, the agent chooses the indicators to query. Technical-persona agents
query technical series (moving averages, turnover, ultra-large order net flow);
fundamental-persona agents query valuation ratios (PE, PB, PS, dividend yield)
when fundamental indicators are enabled. The query window always ends at the
previous close, so the indicator block itself contains no future information.
In the stock/cash and composition scans and in the no-fundamental-information
row of the information experiment, fundamental indicators are switched off, and
fundamental-persona agents work from prices, volumes and holdings only; the
fundamental-data-on cell switches them on. The news experiment uses the
fundamental-data-on cell in both rows, so its contrast isolates the news
channel.

\subsection{Daily Public-Reference Prompt Block}
\label{app:implementation:prompt}

The public reference is announced only through the daily prompt. It appears in
the analysis stage and in all five per-round decision prompts, always as the
current day's single value; the rest of the path is never shown, and the
no-reference arm uses otherwise byte-identical prompts, with the reference block removed. The
following is a translation of the real block from an archived interaction (CGEI, true arm, 2023-06-15); 1{,}016.58 is the day's normalized target level.

\begin{quote}
\footnotesize\ttfamily\raggedright
\#\# External institutional reference\\
- Institutional reference price: 1,016.58\\
- Current price is 1.63\% below the institutional reference price.
\end{quote}

The analysis stage uses the same information with slightly different wording
and an explicit valuation judgement:

\begin{quote}
\footnotesize\ttfamily\raggedright
\#\# External institutional reference information\\
- Institutional reference price: 1,016.58\\
- Institutional valuation judgement: the current price is 1.63\% below the
institutional reference price.
\end{quote}

The flipped-arm block carries the mirrored value and recomputes the comparison
direction; apart from this block, the decision prompt is a fixed
progress-and-rules template shared by both arms.

The templates are originally in Chinese in the released code; the blocks
above are faithful translations. The only difference between the reference
and no-reference arms is the presence of the reference block (and its valuation
line); we verified this in the templates and in archived prompts, where
the no-reference decision prompt begins and ends without any reference
paragraph.

\subsection{Run Inventory}
\label{app:runs}

All reported cells were produced by archived runs. Every reference condition is
run as a true arm and a flipped arm; each no-reference condition has one output
per asset, scored against both targets. Each reported cell corresponds to an
archived run directory that stores, next to its outputs, a snapshot of the
configuration used, so the effective configuration of every reported number is
recoverable. The archived directories contain the multiple independent runs
averaged into each reported cell, including an independently repeated news-on
batch; a few archived runs are pilot or exploratory batches that are not used in
any reported number. The released artifact includes the configuration files and
a condition-to-run mapping that identifies each reported number by condition
family, asset, announced reference and news setting.

\section{Additional Experimental Results}
\label{app:per-asset}

This appendix gives the per-asset numbers behind the aggregate tables. Each
cell is the mean of multiple independent runs per asset and arm, so
differences of a few hundredths in the aggregate should be read as descriptive.
All values are computed from the archived daily close paths with the score of
Appendix~\ref{app:audit}, and the five-asset means reproduce the main-text
tables exactly.

\subsection{Per-Asset Control Effectiveness}
\label{app:per-asset:control}

Table~\ref{tab:app-control-perasset} reports the true-target control
experiment. $\Q_{\mathrm{no}}$ is the realization score of the no-reference
output against the true target, $\Q_{\mathrm{ref}}$ is the same score for the
reference arm's output against the true target, and Gain is their difference.
All ten true-reference comparisons are positive; the flipped-reference
comparisons are positive as well (Appendix~\ref{app:per-asset:fsr}). The two news conditions differ in
both the level of $\Q_{\mathrm{no}}$ (news raises it for three of five assets)
and the size of the gain: under news the reference has less marginal work to do
for four assets and produces the largest gain on TTEI $(+0.403)$, whose
no-reference score collapses to $0.369$. Table~\ref{tab:app-fsr-gain} gives the
flipped-reference counterpart, so the four groups are fully tabulated.

\begin{table}[t]
\centering
\small
\caption{Per-asset control effectiveness on the true target. Both news
conditions include the fundamental data and therefore lie in the
information-injection regime. $\Q_{\mathrm{no}}$ and $\Q_{\mathrm{ref}}$ are the
no-reference and reference realization scores against $\MT$; Gain
$=\Q_{\mathrm{ref}}-\Q_{\mathrm{no}}$ computed from unrounded values. The mean
rows are the aggregates (Table~\ref{tab:control}).}
\label{tab:app-control-perasset}
\setlength{\tabcolsep}{3.6pt}%
\begin{tabular}{@{}lcccccc@{}}
\toprule
& \multicolumn{3}{c}{News off} & \multicolumn{3}{c}{News on} \\
\cmidrule(lr){2-4}\cmidrule(lr){5-7}
Asset & $\Q_{\mathrm{no}}$ & $\Q_{\mathrm{ref}}$ & Gain &
$\Q_{\mathrm{no}}$ & $\Q_{\mathrm{ref}}$ & Gain \\
\midrule
CGEI & 0.5344 & 0.9132 & $+0.3788$ & 0.7085 & 0.9055 & $+0.1970$ \\
EREI & 0.6205 & 0.9259 & $+0.3055$ & 0.7318 & 0.8548 & $+0.1230$ \\
FSEI & 0.7134 & 0.8785 & $+0.1650$ & 0.6905 & 0.7529 & $+0.0623$ \\
REEI & 0.6541 & 0.9382 & $+0.2841$ & 0.7368 & 0.8686 & $+0.1318$ \\
TTEI & 0.5351 & 0.9340 & $+0.3989$ & 0.3688 & 0.7716 & $+0.4028$ \\
\midrule
Mean & 0.6115 & 0.9180 & $+0.3065$ & 0.6473 & 0.8307 & $+0.1834$ \\
\bottomrule
\end{tabular}
\end{table}

\subsection{Per-Asset Flipped-Arm Gain}
\label{app:per-asset:fsr}

Table~\ref{tab:app-fsr-gain} reports the flipped-arm counterpart. Here $\Q_{\mathrm{no}}$ is the no-reference output scored against the
\emph{flipped} target $\MF$, $\mathrm{FSR}$ is the reference arm's flipped-arm
score against $\MF$, and the FSR gain is their difference. The reference raises
the flipped-arm score for every asset under both news conditions; the FSR gains
are larger on aggregate than the true-target gains in the same cell, because the
no-reference output is far from the flipped path (e.g.\ TTEI: $0.191\to0.842$
without news). This follows from the ceiling on the gain, and it is why $\R$ is read
as realization \emph{under the announced reference} rather than as a general
capability of the model.

\begin{table}[t]
\centering
\footnotesize
\caption{Per-asset flipped-arm gain, in the information-injection regime.
$\Q_{\mathrm{no}}$ is the no-reference output scored against $\MF$;
$\mathrm{FSR}$ is the reference arm's flipped-arm score against $\MF$; FSR gain
is their difference (computed from unrounded values). Means are over the five
assets.}
\label{tab:app-fsr-gain}
\setlength{\tabcolsep}{2.5pt}%
\begin{tabular}{@{}lcccccc@{}}
\toprule
& \multicolumn{3}{c}{News off} & \multicolumn{3}{c}{News on} \\
\cmidrule(lr){2-4}\cmidrule(lr){5-7}
Asset & $\Q_{\mathrm{no}}$ & FSR & FSR gain &
$\Q_{\mathrm{no}}$ & FSR & FSR gain \\
\midrule
CGEI & 0.6382 & 0.8896 & $+0.2514$ & 0.3887 & 0.7608 & $+0.3722$ \\
EREI & 0.5989 & 0.9157 & $+0.3167$ & 0.3777 & 0.8325 & $+0.4548$ \\
FSEI & 0.3510 & 0.8392 & $+0.4882$ & 0.3059 & 0.5168 & $+0.2109$ \\
REEI & 0.5645 & 0.9330 & $+0.3684$ & 0.4680 & 0.9260 & $+0.4580$ \\
TTEI & 0.1911 & 0.8421 & $+0.6510$ & 0.7263 & 0.8739 & $+0.1476$ \\
\midrule
Mean & 0.4688 & 0.8839 & $+0.4152$ & 0.4533 & 0.7820 & $+0.3287$ \\
\bottomrule
\end{tabular}
\end{table}

\subsection{Per-Asset Realization Across Conditions}
\label{app:per-asset:conditions}

Table~\ref{tab:app-perasset-conditions} gives the per-asset TSR and FSR for
the reference-only scans and the two information-injection cells, with each
cell shown as ``TSR / FSR''. The per-condition means reproduce
Tables~\ref{tab:c1}, \ref{tab:c2}, \ref{tab:news} and \ref{tab:c3}. The scans are not monotone in aggregate, nor in nine of the ten
asset-by-scan combinations; for example the 90/10 allocation leaves REEI at
$0.2202/0.6096$ while CGEI stays at $0.9169/0.8955$. The technical-only
composition collapses in every asset, and the reduction is concentrated on the
TSR side. The composition 3/7 cell is the shared baseline reused throughout the
paper and is identical to the stock/cash 70/30 cell.

\begin{table*}[t]
\centering
\scriptsize
\caption{Per-asset TSR / FSR by condition. Each cell is
$\mathrm{TSR}/\mathrm{FSR}$ for that asset. Reference-only rows use that
condition's true and flipped arms; the last two rows are the
information-injection cells of the fundamental-data and news experiments
(their news-off row is the fundamental-data-on cell). The composition
3/7 row is the shared baseline reused by the other experiments; it is the same
cell as stock/cash 70/30 and is listed here for completeness.}
\label{tab:app-perasset-conditions}
\setlength{\tabcolsep}{6pt}%
\begin{tabular}{@{}lccccc@{}}
\toprule
Condition & CGEI & EREI & FSEI & REEI & TTEI \\
\midrule
Stock/cash 30/70 & 0.8072/0.3702 & 0.7532/0.7143 & 0.8076/0.7733 & 0.9061/0.8886 & 0.6271/0.8826 \\
Stock/cash 50/50 & 0.8524/0.6547 & 0.8253/0.8038 & 0.8810/0.8119 & 0.9413/0.9180 & 0.7638/0.9227 \\
Stock/cash 70/30 & 0.9316/0.8688 & 0.8947/0.9119 & 0.8850/0.8919 & 0.8992/0.8975 & 0.8481/0.9148 \\
Stock/cash 90/10 & 0.9169/0.8955 & 0.6791/0.6852 & 0.8005/0.9249 & 0.2202/0.6096 & 0.8282/0.8267 \\
\midrule
Composition 0/10 & 0.8955/0.8786 & 0.8704/0.8876 & 0.9115/0.9148 & 0.8861/0.9044 & 0.8507/0.8732 \\
Composition 3/7 (shared) & 0.9316/0.8688 & 0.8947/0.9119 & 0.8850/0.8919 & 0.8992/0.8975 & 0.8481/0.9148 \\
Composition 5/5 & 0.8850/0.9236 & 0.8827/0.8039 & 0.9200/0.8794 & 0.6575/0.4899 & 0.8344/0.8574 \\
Composition 10/0 & 0.3463/0.4926 & 0.1540/0.4965 & 0.1920/0.4136 & 0.1803/0.4360 & 0.5095/0.3602 \\
\midrule
Fundamental data on, news off & 0.9132/0.8896 & 0.9259/0.9157 & 0.8785/0.8392 & 0.9382/0.9330 & 0.9340/0.8421 \\
Fundamental data on, news on & 0.9055/0.7608 & 0.8548/0.8325 & 0.7529/0.5168 & 0.8686/0.9260 & 0.7716/0.8739 \\
\bottomrule
\end{tabular}
\end{table*}

\subsection{Per-Asset No-Reference Baseline}
\label{app:per-asset:noref}

Table~\ref{tab:app-noref} reports the auxiliary no-reference check used in the
main text. The same no-reference output is scored against the true target
($\Q_{\mathrm{T}}$) and against the flipped target ($\Q_{\mathrm{F}}$);
$\D_{\mathrm{no}}=\Q_{\mathrm{T}}-\Q_{\mathrm{F}}$. This setting lies outside
the two-regime framework and is a consistency check only. It shows that news
raises the true-over-flipped preference in four of five assets, the exception
being TTEI, whose no-reference output moves towards the flipped path once news
is on; the aggregate rise in $\D_{\mathrm{no}}$ is therefore the net of
opposite asset-level movements. The no-reference output used in the table is the archived daily close
path.

\begin{table}[t]
\centering
\small
\caption{Per-asset no-reference baseline for the news experiment. The same
no-reference output is scored against $\MT$ ($\Q_{\mathrm{T}}$) and $\MF$
($\Q_{\mathrm{F}}$); $\D_{\mathrm{no}}=\Q_{\mathrm{T}}-\Q_{\mathrm{F}}$.
Means are over the five assets.}
\label{tab:app-noref}
\setlength{\tabcolsep}{3.6pt}%
\begin{tabular}{@{}lcccccc@{}}
\toprule
& \multicolumn{3}{c}{News off} & \multicolumn{3}{c}{News on} \\
\cmidrule(lr){2-4}\cmidrule(lr){5-7}
Asset & $\Q_{\mathrm{T}}$ & $\Q_{\mathrm{F}}$ & $\D_{\mathrm{no}}$ &
$\Q_{\mathrm{T}}$ & $\Q_{\mathrm{F}}$ & $\D_{\mathrm{no}}$ \\
\midrule
CGEI & 0.5344 & 0.6382 & $-0.1038$ & 0.7085 & 0.3887 & $+0.3198$ \\
EREI & 0.6205 & 0.5989 & $+0.0215$ & 0.7318 & 0.3777 & $+0.3541$ \\
FSEI & 0.7134 & 0.3510 & $+0.3624$ & 0.6905 & 0.3059 & $+0.3846$ \\
REEI & 0.6541 & 0.5645 & $+0.0896$ & 0.7368 & 0.4680 & $+0.2688$ \\
TTEI & 0.5351 & 0.1911 & $+0.3440$ & 0.3688 & 0.7263 & $-0.3575$ \\
\midrule
Mean & 0.6115 & 0.4688 & $+0.1427$ & 0.6473 & 0.4533 & $+0.1940$ \\
\bottomrule
\end{tabular}
\end{table}

\section{Robustness and Calibration Checks}
\label{app:validation}

This section gives the construction details behind the design-validation
checks summarized in the main text.

\subsection{Null Baselines}
\label{app:validation:null}

Three zero-information path constructions are scored with the same metric
against the five true targets, on the same 51 days.

\begin{itemize}[leftmargin=*,itemsep=1pt,topsep=2pt]
  \item \textbf{Constant:} $P_t=1000$ for all $t$. It has no shape and no
  dynamics; its Level is set by the average distance of the target from the
  normalization level.
  \item \textbf{Linear:} the straight line between the target's first and last
  level. It uses both endpoints, i.e.\ it is a strong strawman that
  knows where the path starts and ends.
  \item \textbf{Matched-volatility random walk:} a zero-drift Gaussian random
  walk on log prices, $P_0=1000$, with daily volatility equal to the target's
  own $\operatorname{std}(\Delta\log M)$; 5{,}000 seeded paths per asset. We
  report the mean score over paths and also use the path distribution to
  locate the model's score.
\end{itemize}

As a supplemental check we also score \emph{shuffled returns}: the target's own
daily log returns are randomly permuted and cumulated, preserving the return
multiset and the total return but destroying the ordering. Table~\ref{tab:app-null}
reports the results together with the model's no-reference and reference
scores. Three readings follow. (i) The reference arm lies above all four nulls
and above the 95th percentile of the matched-volatility random walk in five of
five assets without news and three of five with news; the random-walk null has
a wide upper tail (5th--95th percentile of path score $0.318$--$0.804$), so the
claim is at the distribution level, not that no random path can reach $0.9$.
(ii) The no-reference score is essentially the constant-path score
($0.6115$ vs $0.6104$ without news) and below the shuffled-return strawman
($0.6821$): without the announced reference the output is statistically
indistinguishable from a zero-information path. (iii) The linear null still
lags the reference by $0.097$ without news, which shows that the reference
effect is not merely endpoint knowledge.

\begin{table}[t]
\centering
\small
\caption{Zero-information nulls and model scores (means over the five true
targets). Percentile columns locate the model score in the per-asset
distribution of the matched-volatility random-walk null. ``--'' means not
applicable.}
\label{tab:app-null}
\setlength{\tabcolsep}{3.6pt}%
\begin{tabular}{@{}lcccc@{}}
\toprule
Path / model & $\bar{\Q}$ & Level & Shape & Percentile \\
\midrule
Constant & 0.6104 & 0.7207 & 0.5000 & 56 \\
Linear interpolation & 0.8210 & 0.8121 & 0.8299 & 96 \\
Matched-vol. RW & 0.5650 & 0.6306 & 0.4994 & -- \\
Shuffled returns & 0.6821 & 0.7308 & 0.6333 & 69 \\
\midrule
No-reference, news off & 0.6115 & 0.5979 & 0.6251 & 58 \\
Reference, news off & 0.9180 & 0.8906 & 0.9453 & 100 \\
No-reference, news on & 0.6473 & 0.6489 & 0.6456 & 67 \\
Reference, news on & 0.8307 & 0.8008 & 0.8605 & 94 \\
\bottomrule
\end{tabular}
\end{table}

\subsection{Score Sensitivity}
\label{app:validation:sensitivity}

We recompute $\R$ and $\B$ on the reference-only baseline cell and $\D$ on
the news experiment under five definitions of $\Q$ on the same frozen runs:
\begin{itemize}[leftmargin=*,itemsep=1pt,topsep=2pt]
  \item \textbf{Paper $\Q$:} $\tfrac12\mathrm{Level}+\tfrac12\mathrm{Shape}$
  with $\sigma_M$ equal to the arm's own target volatility.
  \item \textbf{Common $\sigma$:} the Level denominator uses the true target's
  volatility for both arms, removing the true/flipped volatility asymmetry.
  \item \textbf{Shape on returns:} Shape is the Spearman correlation of daily
  log returns of $P$ and $M$ instead of levels.
  \item \textbf{Shape only:} $\Q=\mathrm{Shape}$.
  \item \textbf{Level only:} $\Q=\mathrm{Level}$.
\end{itemize}

Table~\ref{tab:app-sensitivity} reports the results. The 70/30 baseline has a
low $\B$ under every definition, $\R$ stays in the band $0.760$--$0.924$, and
$\D$ without news stays positive $(+0.022$--$+0.046)$. The news increment is
positive under the paper's $\Q$, the common-$\sigma$ normalization and the
shape-only definition, and negative under shape-on-returns and level-only; the
variants are reported here for transparency, and the paper's score is used
throughout the main text. Under shape-on-returns the absolute size of $\R$ falls
to $0.760$.

\begin{table*}[t]
\centering
\footnotesize
\caption{Score sensitivity. $\R$ and $\B$ are the reference-only 70/30 cell;
$\D$ is the news experiment. ``News effect'' is $\D_{\mathrm{on}}-\D_{\mathrm{off}}$
(or its sign); these are the values discussed in
Section~\ref{sec:results:validation}.}
\label{tab:app-sensitivity}
\setlength{\tabcolsep}{4pt}%
\begin{tabular}{@{}lcccccc@{}}
\toprule
$\Q$ definition & $\R$ & $\B$ & $\D$ off & $\D$ on & News eff. & Sign \\
\midrule
Paper $\Q$ & 0.8943 & 0.0418 & $+0.0341$ & $+0.0487$ & $+0.0146$ & $+$ \\
Common $\sigma$ & 0.8938 & 0.0444 & $+0.0342$ & $+0.0503$ & $+0.0161$ & $+$ \\
Shape on returns & 0.7602 & 0.0476 & $+0.0386$ & $-0.0029$ & $-0.0415$ & $-$ \\
Shape only & 0.9235 & 0.0356 & $+0.0218$ & $+0.0828$ & $+0.0610$ & $+$ \\
Level only & 0.8652 & 0.0506 & $+0.0463$ & $+0.0146$ & $-0.0318$ & $-$ \\
\bottomrule
\end{tabular}
\end{table*}

\subsection{Asset-Level Bootstrap}
\label{app:validation:bootstrap}

Because the aggregate scores are means over five independent asset runs, we
resample the five asset pairs with replacement (20{,}000 draws) and report the
percentile 95\% interval. Resampling the five assets measures how much the
aggregate depends on which assets happen to be in the suite; it does not
capture run-to-run sampling noise, which is averaged out in the reported means
and additionally checked against an independent repeated batch. With $n=5$ the bootstrap distribution is coarse (at most
$\binom{9}{5}=126$ distinct weightings, or $5^5=3125$ ordered draws) and the interval should be read as a
descriptive range, not a calibrated confidence interval. Table~\ref{tab:app-bootstrap}
reports the key intervals. The reference-effect gains have intervals away from
zero in both news conditions. The no-reference $\D$ intervals include zero
under both news conditions, consistent with the interpretation that the
no-reference output is not supposed to track any target. The news-on $\D$
interval includes zero, which is the quantitative counterpart of the
metric-dependence noted above.

\begin{table}[t]
\centering
\small
\caption{Asset-level bootstrap intervals (percentile, 20{,}000 draws over the
five asset pairs; intervals are discrete and descriptive). $\D$ is
$\mathrm{TSR}-\mathrm{FSR}$, and the gains are reported separately for the true and flipped references under each news condition.}
\label{tab:app-bootstrap}
\setlength{\tabcolsep}{3.6pt}%
\begin{tabular}{@{}lcc@{}}
\toprule
Estimand & Point & 95\% interval \\
\midrule
Gain (true), news off & 0.3065 & $[0.2316,\ 0.3722]$ \\
Gain (flipped), news off & 0.4152 & $[0.3009,\ 0.5516]$ \\
Gain (true), news on & 0.1834 & $[0.1005,\ 0.2944]$ \\
Gain (flipped), news on & 0.3287 & $[0.2178,\ 0.4395]$ \\
\midrule
$\R$, reference-only 70/30 & 0.8943 & $[0.8866,\ 0.9011]$ \\
$\B$, reference-only 70/30 & 0.0418 & $[0.0089,\ 0.0589]$ \\
\midrule
$\D$, news off & 0.0341 & $[0.0109,\ 0.0646]$ \\
$\D$, news on & 0.0487 & $[-0.0594,\ 0.1568]$ \\
$\D_{\mathrm{no}}$, news off & 0.1427 & $[-0.0150,\ 0.3005]$ \\
$\D_{\mathrm{no}}$, news on & 0.1940 & $[-0.0865,\ 0.3595]$ \\
\midrule
News effect on $\D$ & $+0.0146$ & $[-0.1049,\ 0.1298]$ \\
\bottomrule
\end{tabular}
\end{table}

\subsection{Replicate Batch}
\label{app:validation:replicate}

Two independent batches of the news-on condition were run with identical
configuration: the reported batch and an earlier repeat batch. Table~\ref{tab:app-replicate}
gives the per-asset $\D$ for both. The aggregate changes from $0.0487$ to
$0.0632$ and TSR/FSR from $0.8307/0.7820$ to $0.8552/0.7920$, i.e.\ the two
batches agree on the sign of the aggregate effect but not on its size. At the
asset level only CGEI changes the sign of $\D$ ($+0.1447$ vs $-0.0022$); the
other four assets preserve their sign. Differences of a few hundredths in
$\D$ are therefore within run-to-run variability, and the reported cell is one
draw from that distribution. This is why the paper treats all reported means as
descriptive effect sizes (Section~\ref{sec:discussion}).

\begin{table}[t]
\centering
\small
\caption{Replicate batch of the news-on condition: per-asset
$\D=\mathrm{TSR}-\mathrm{FSR}$. ``Reported'' is the batch used in the tables;
``repeat'' is the earlier independent batch; the last column is repeat minus
reported.}
\label{tab:app-replicate}
\begin{tabular}{@{}lccc@{}}
\toprule
Asset & Reported $\D$ & Repeat $\D$ & Difference \\
\midrule
CGEI & $+0.1447$ & $-0.0022$ & $-0.1469$ \\
EREI & $+0.0223$ & $+0.0214$ & $-0.0009$ \\
FSEI & $+0.2361$ & $+0.3451$ & $+0.1090$ \\
REEI & $-0.0574$ & $-0.0333$ & $+0.0242$ \\
TTEI & $-0.1023$ & $-0.0148$ & $+0.0875$ \\
\midrule
Mean & $+0.0487$ & $+0.0632$ & $+0.0146$ \\
\bottomrule
\end{tabular}
\end{table}

\section{Reproducibility Statement}
\label{app:repro}

\paragraph{Code and artifacts.} All runs use the TwinMarket codebase at
commit \texttt{df3fdcb42ee7d300378eb88a}\allowbreak
\texttt{51858195fac20aa5}. Each archived run stores a configuration snapshot
next to its outputs, so the effective configuration of every reported run is
recoverable. The release will include the code snapshot, the configuration
files, the prompts, and the analysis code that recomputes every reported score
from the archived daily close paths.

\paragraph{Environment.} All runs were executed on Windows 11 with Python
3.12.4 in a virtual environment and the dependencies pinned in
\texttt{requirements.txt} (notably \texttt{openai}, \texttt{pandas},
\texttt{numpy}, \texttt{faiss-cpu}, \texttt{networkx}, \texttt{PyYAML},
\texttt{tenacity}, \texttt{tqdm}; \texttt{scipy} for the evaluation scripts).
No GPU is required. The agent model is the hosted DeepSeek-Chat API
(\texttt{deepseek-chat}, \url{https://api.deepseek.com}); the model is a
provider-side rolling alias and no version snapshot was recorded, which is an
explicit reproducibility limitation. The paper was compiled with pdf\LaTeX{}
(TeX Live 2025).

\paragraph{Reproducing one condition.} A cell is reproduced from the released
code and configuration files: the run directory is prepared from the archived
inputs, the simulator is launched over the 51-day window, and the archived daily
close path is scored with the metric of Appendix~\ref{app:audit}. The released
condition-to-run mapping identifies the archived run behind every reported
number, and the release includes the configuration snapshots that were used. Conditions that differ in one factor use the same pipeline with the
corresponding switch (prefix, composition, initial allocation, news on/off,
technical and fundamental indicators on/off, reference on/off).

\paragraph{Data availability.} The release contains (i) the composite index
series and constituent weights of the five evaluated assets (and of the ten
platform sectors), (ii) the daily news stream and the retrieval index metadata,
(iii) the synthetic investor database used to generate the ten-agent runs,
(iv) the archived run directories, including per-call prompt traces, model
responses, orders, fills, daily close paths, and configuration snapshots, and
(v) the analysis scripts. The composite indices are locally constructed
research series; we release the series and the constituent weights, but not
constituent-level price data from the commercial source. The news items are
public financial news collected in the local store. Code, data, configs and
result files will be released publicly upon acceptance.

\paragraph{Scope of the evidence.} Each reported condition averages multiple independent runs per asset and arm, and an
independently repeated news-on batch provides an additional stability check
(Appendix~\ref{app:validation:replicate}). The main tables report condition
means; asset-level bootstrap intervals are given in
Appendix~\ref{app:validation:bootstrap}, and the prompts are archived verbatim
for every call in the prompt traces. The run mapping is included in the release.

\section{Use of Large Language Models}
\label{app:llm}

Large language models were used in two distinct roles. First, they are the
object of study: the simulated investors are DeepSeek-Chat agents, and their
prompts, responses and orders are archived as described above. Second, language
models were used as assistants during the preparation of this paper and the
codebase, for language editing, code writing and code review. The authors
specified the research questions, the experimental design, the metric and the
analysis, verified every reported number against the archived runs, and take
full responsibility for the content of the paper and for the released code. No
LLM is an author or contributor, and no LLM-generated text is presented as
evidence or as a finding. The study itself involves no human subjects and no
real trading; its ethics considerations are stated in the Ethics Statement of
Appendix~\ref{sec:ethics}.

%% file: appendix_extra.tex
\section{Additional Figures and Tables}
\label{app:figs}

This appendix collects additional figures referenced from the main text.

\begin{figure}[t]
  \centering
  \includegraphics[width=\columnwidth]{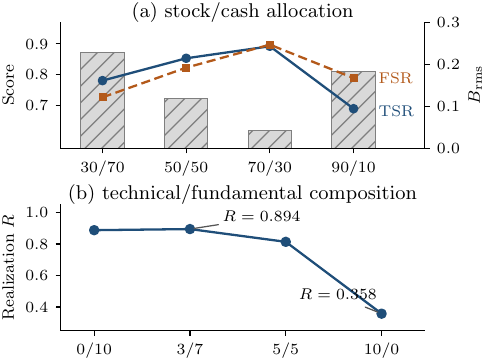}
  \caption{Structural scans in the reference-only regime. (a) Stock/cash allocation versus TSR and FSR (lines, left axis) and structural bias $\B_{\mathrm{rms}}$ (bars, right axis); bias is lowest at 70/30 ($\B_{\mathrm{rms}}=0.042$). (b) Technical-agent share versus realization capability $\R$; an intermediate share (3:7) realizes best ($\R=0.894$), whereas a technical-only composition collapses ($\R=0.358$).}
  \label{fig:scan}
\end{figure}

\begin{figure}[t]
  \centering
  \includegraphics[width=0.95\columnwidth]{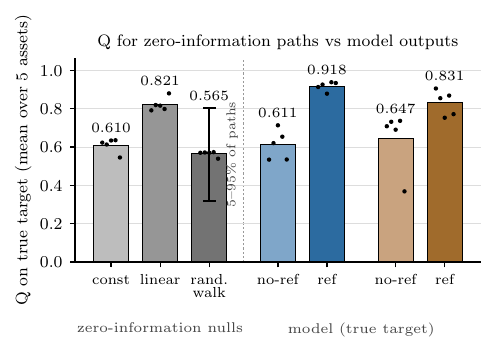}
  \caption{Null baselines for the realization score. Constant path at the normalization level, straight-line interpolation between the first and last target level, a random walk matched to the target's volatility, and the no-reference and reference arms, all scored with the paper's $\Q$ (mean over the five true targets). Dots are the five per-asset values; the whisker on the random-walk bar is the 5th--95th percentile over 5{,}000 seeded paths (averaged over assets), not an error bar of the mean. The reference arm under news is only $0.010$ above the linear-interpolation null.}
  \label{fig:null}
\end{figure}

Figure~\ref{fig:scan} shows the full structural scans summarized in the main text, Figure~\ref{fig:null} reports the null baselines and Figure~\ref{fig:sens} the score-sensitivity check of Section~\ref{sec:results:validation}.

\begin{table}[t]
\centering
\footnotesize
\setlength{\tabcolsep}{2.5pt}
\caption{News experiment. Both conditions include the fundamental data and therefore lie in the information-injection regime: TSR, FSR, and $\D_{\mathrm{ref}}$ are computed under the public reference, and $\D_{\mathrm{ref}}=\mathrm{TSR}-\mathrm{FSR}$ is the interpretable quantity. The last three columns report an independent no-reference baseline that is outside the two-regime framework: the same no-reference output is scored against the true target ($\Q_{\mathrm{T}}$) and the flipped target ($\Q_{\mathrm{F}}$), and $\D_{\mathrm{no}}=\Q_{\mathrm{T}}-\Q_{\mathrm{F}}$.}
\label{tab:news}
\begin{tabular}{@{}lcccccc@{}}
\toprule
News & TSR & FSR & $\D_{\mathrm{ref}}$ & $\Q_{\mathrm{T}}$ & $\Q_{\mathrm{F}}$ & $\D_{\mathrm{no}}$ \\
\midrule
Off & 0.9180 & 0.8839 & $+0.0341$ & 0.6115 & 0.4688 & $+0.1427$ \\
On  & 0.8307 & 0.7820 & $+0.0487$ & 0.6473 & 0.4533 & $+0.1940$ \\
\bottomrule
\end{tabular}
\end{table}

\begin{figure}[t]
  \centering
  \includegraphics[width=0.95\columnwidth]{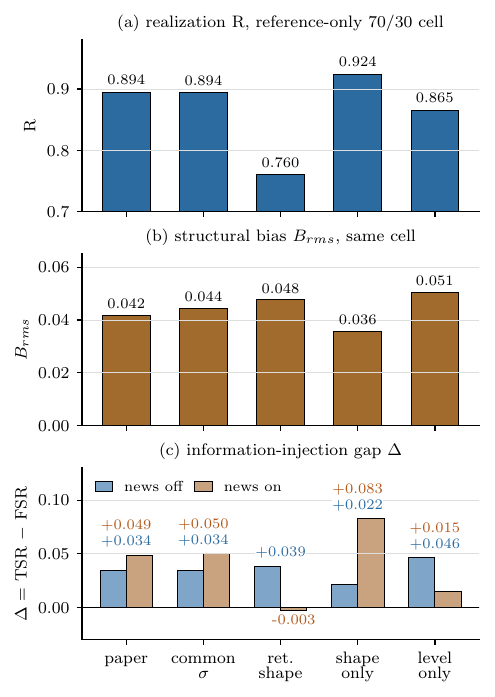}
  \caption{Score sensitivity. $\R$ and $\B$ on the 70/30, 3:7, news-off cell and $\D$ on the news experiment, under the five definitions of the realization score in Table~\ref{tab:app-sensitivity}. The news increment changes sign under shape-on-returns and level-only.}
  \label{fig:sens}

  \vspace{4pt}
  \captionof{table}{Fundamental-information experiment. The first row is reference-only and
serves as the symmetry check; the second row is information-injection. Absolute
TSR/FSR are recorded for transparency, and only $\bar{\D}$ in the injection regime
is interpreted; the two rows are not compared across regimes.}
  \label{tab:c3}
  \footnotesize
  \setlength{\tabcolsep}{4pt}
  \begin{tabular}{@{}p{0.27\columnwidth}p{0.22\columnwidth}ccc@{}}
  \toprule
  Condition & Regime & TSR & FSR & $\bar{\D}$ \\
  \midrule
  No fundamental info. & reference-only & 0.8917 & 0.8970 & $-0.0052$ \\
  Fundamental info. & information-injection & 0.9180 & 0.8839 & $+0.0341$ \\
  \bottomrule
  \end{tabular}
\end{figure}

\section{Ethics Statement}
\label{sec:ethics}

This study uses publicly available historical index data and synthetic LLM agents; it involves no human subjects and no real trading. The simulated market is a research environment, and nothing in this paper constitutes investment advice or a recommendation to deploy the system in real markets. LLM-generated agent behavior may reflect biases present in the model or its pretraining data; the evaluation framework reports aggregate behavior and is not intended to certify any individual model output. The framework could in principle be used to study manipulative information channels; we use it only with public news and fundamental data, and we see no additional dual-use risk beyond that of standard market-simulation research.